\documentclass{article}

\usepackage{hyperref}
\usepackage{url}
\usepackage{booktabs}
\usepackage{amsfonts}
\usepackage{nicefrac}
\usepackage{microtype}
\usepackage{graphicx}
\usepackage{color}
\usepackage{amsmath}

\usepackage[a4paper, total={6.2in, 9.3in}]{geometry}

\usepackage{caption}
\usepackage{subcaption}
\usepackage{verbatim}
\usepackage{multirow}

\usepackage{multirow}
\usepackage{authblk}
\usepackage{rotating}

\usepackage{authblk}

\title{\huge{Attention Incorporate Network: A network can adapt various data size}}

\author{Liangbo He}
\author{Hao Sun}
\affil{
    \texttt{\normalsize \{heliangbo, sh759811581\}@tsinghua.edu.cn}
}
\affil{Tsinghua University\\Hai Dian, Beijing, China}

\date{}

\begin{document}

\maketitle
\begin{abstract}
  In traditional neural networks for image processing, the inputs of the neural networks should be the same size such as 224$\times$224$\times$3. But how can we train the neural net model with different input size?
  A common way to do is image deformation which accompany a problem of information loss (e.g. image crop or wrap). Sequence model(RNN, LSTM, etc.) can accept different size of input like text and audio. But one disadvantage for sequence model is that the previous information will become more fragmentary during the transfer in time step, it will make the network hard to train especially for long sequential data. In this paper we propose a new network structure called Attention Incorporate Network(AIN). It solve the problem of different size of inputs including: images, text, audio, and extract the key features of the inputs by attention mechanism, pay different attention depends on the importance of the features not rely on the data size. Experimentally, AIN achieve a higher accuracy, better convergence comparing to the same size of other network structure.
  
\end{abstract}

\section{Introduction}
\begin{figure}[ht]
  \begin{minipage}[t]{0.5\linewidth} 
  \centering
  \includegraphics[width=1.5in]{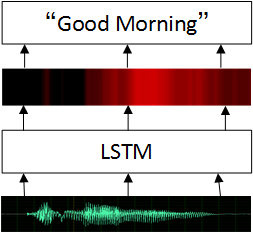}
  \caption{LSTM structure.}
  \label{fig:pic1}
  \end{minipage}%
  \begin{minipage}[t]{0.5\linewidth} 
  \centering
  \includegraphics[width=1.5in]{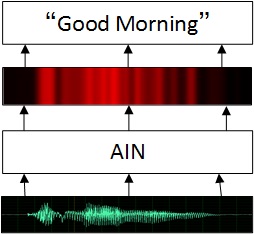}
  \caption{AIN structure.}
  \label{fig:pic2}
  \end{minipage}%
\end{figure}

Human use the attention mechanism to recognize the world, pay different attention on different region of a image. The mixed nature of attention has been studied in the previous literature~\cite{Zhao2017Diversified,Itti2001Computational,Shin2016Lexicon,Wang2017Residual}. Attention-based network is widely used in  sequence model to process text or audio~\cite{Mnih2014Recurrent}. When we were building a trigger word detection application in speech recognition problem, the model were firstly trained based by LSTM~\cite{Hochreiter1997Long}. One problem confused us most is that the elements in trigger word didn't share the same weights. For example, if the trigger word is "\emph{good morning}", "good" and "morning" should have the exactly the same importance for the whole phrase. But LSTM will prefer to give a higher importance to the word which is closer to the end which means "morning" is more likely to be activate than "good"(Fig.~\ref{fig:pic1}). We built a network based on AIN(Fig.~\ref{fig:pic2}) to solve this problem and made a success in the real world application.

\begin{figure}[ht]
  \centering
  \includegraphics[width=1.0\linewidth]{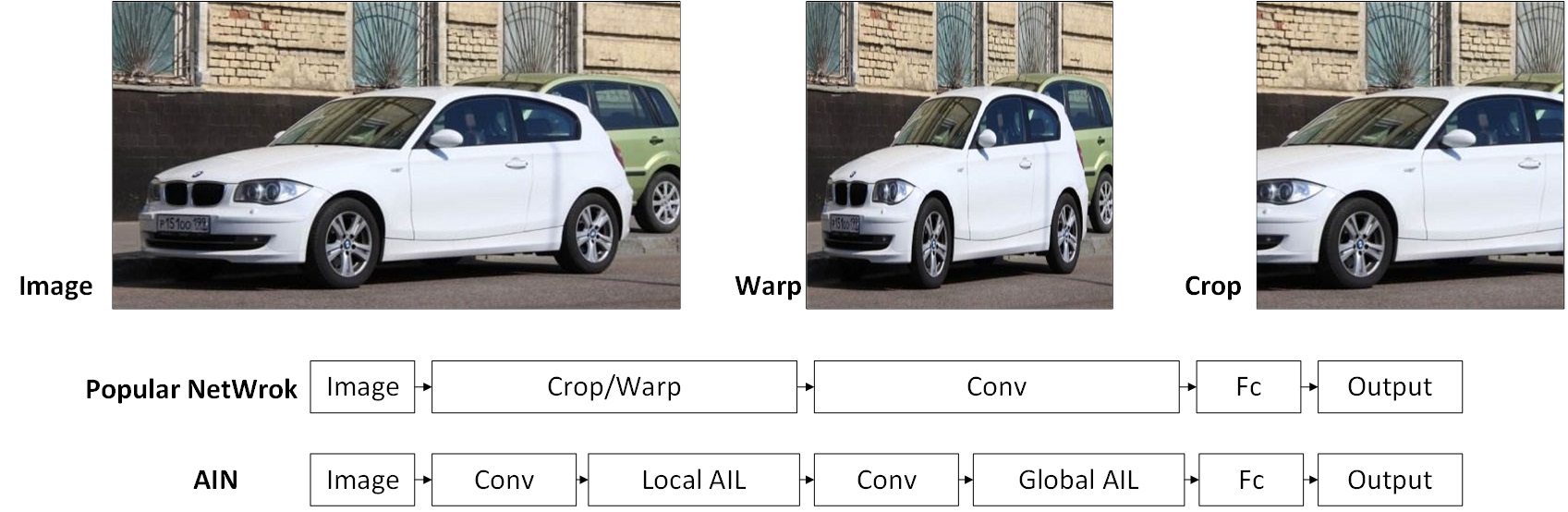}
  \caption{Top:croping or wraping into a fixed size.\\ Bottom:popular network and AIN structure.}
  \label{fig:pic3}
\end{figure}

Inspired by our previous work and the attention mechanism in related literature\cite{Zhao2017Diversified,Itti2001Computational,Shin2016Lexicon,Wang2017Residual}. In this paper we propose Attention Incorporate Network(AIN), rather than sequential data processing, AIN have a huge advantage in image classification problem. Recent advances of image
classification focus on a "very deep" network structure, from AlexNet~\cite{Krizhevsky2012ImageNet} to DenseNet~\cite{Huang2016Densely} the precision become higher and higher, but some problems are still remain to be solved, our model exhibits following appealing properties to solve these problem:

\textbf{(1)}Neural Network come from human neurons, human can recognize the object easily no matter what the image shape is. But all kinds of popular network like AlexNet~\cite{Krizhevsky2012ImageNet},VGG~\cite{Simonyan2014Very}, Inception~\cite{Szegedy2016Inception}and ResNet~\cite{He2016Deep} couldn't process the image in different size. An engineering solution is to scale the image into a same shape, which isn't an aesthetic way due to human neurons won't scale the object image while recognizing it. The back logic in human neuron should be attention mechanism. Our network structure AIN training feed-forward is just like human neurons recognize the object: combine the attention matrix and image together and give a label after incorporate all the key information(Fig.~\ref{fig:pic3}).
The input image can be arbitrary size due to the AIN feed-forward operation which gives a more natural solution than traditional image scaling(Fig.~\ref{fig:pic3}).

\textbf{(2)}In modern convolutional network structure, image size will decrease while convolution layer get deeper. The image information will be transferred into the image channels. Pooling and Convolution with stride are the well known layers to reduces the image size. AlexNet~\cite{Krizhevsky2012ImageNet} used maxpooling, Resnet~\cite{He2016Deep} used Conv1$\times$1 with stride of 2. Information will remain only \nicefrac{1}{s$^2$}~(s=stride or pooling size) during feed-forward the transition layer. In AIN we used \emph{attention incorporate layer}(AIL) to extract the key features of the image without information loss. Experimentally, AIN achieved a lower error comparing to pooling or covolution with stride in the same network structure.
\section{Related Work}
The field of attention is one of the oldest in psychology. Evidence from human perception process\cite{Mnih2014Recurrent} show the important of the attention mechanism which uses top information to guide bottom-up feed-forward process\cite{Wang2017Residual}. Attention was widely used in many applications: Recommendation system~\cite{Vinh2018Attention}, Image classification~\cite{Asada2014Deep,Zhao2017Diversified}, Image caption generation~\cite{Xu2015Show}, Machine translation~\cite{Luong2015Effective,Firat2016Multi}, Speech recognition~\cite{Chorowski2015Attention,Bahdanau2016End} and gave us some good result.

In Image classification, people tried some method to solve the arbitrary input size.
He, K. et. al~\cite{He2015Spatial}. used \emph{spatial pyramid pooling}(SPP) to remove the fixed-size constraint of the network. Pooling size and stride depends on the input data to get a fixed output at the last convolutional layer. The back logic in their network is artificially divide image into fixed numbers of region, which is totally different from human cognitive system. Another disadvantage of SPP is that pooling layer lost lot of information while training feed-forward especially for those \emph{"very deep"} networks.

Pooling reduces the size of the hidden layers so quickly, stacks of back-to-back convolutional layers are needed to build really deep networks. Zeiler et. al~\cite{Zeiler2013Stochastic} introduce Stochastic Pooling where the act of picking the maximum value in each pooling region is replaced by a form of size-biased sampling. Graham et. al~\cite{Graham2014Fractional} introduce fractional max-pooling (FMP). The idea of FMP is to reduce the spatial size of the image by a factor of $\alpha$ with 1 < $\alpha$ < 2. Their solutions reduce the information loss of the pooling layer but still left the problem behind. 
\section{Attention Incorporate Network}
\label{headings}

Our Attention Incorporate Network constructed by stacking multiple \emph{local attention incorporate layer}(LAIL) and a \emph{global attention incorporate layer}(GAIL) before the FC-layer. The output of LAIN can be any size
depend on the input, and the output of GAIN will be the fixed size design by the network structure to ensure that no size conflict between the GAIN and FC-layer. And for both GAIL and LAIL share exactly the same mathematical calculation.

\subsection{Attention incorporate layer}

\begin{figure}[h]
  \centering
  \includegraphics[width=2.5in]{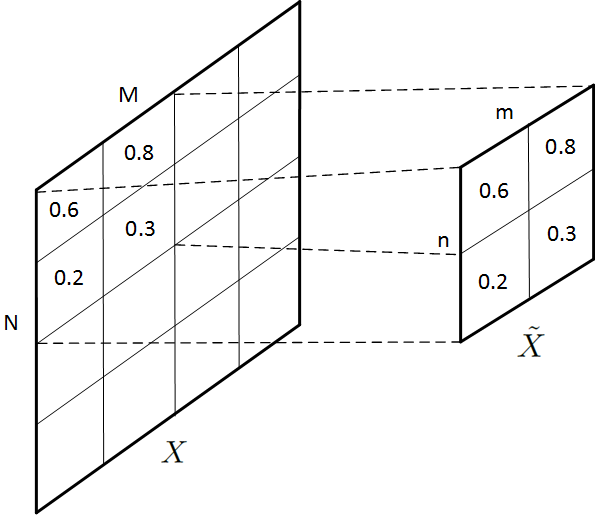}
  \caption{Select Window Operation.}
  \label{fig:pic20}
\end{figure}

\paragraph{Forward Pass.}Traditional convolutional feed-forward networks connect the  output of the $\ell ^{th}$ layer as input to the
$(\ell+1)^{th}$ layer~\cite{Krizhevsky2012ImageNet}. The transition layer in AlexNet~\cite{Krizhevsky2012ImageNet}, Resnet~\cite{He2016Deep} fellow the equation(1)(2):
\begin{equation}
 X_{\ell+1} = MaxPooling(X_{\ell}) 
\end{equation}

\begin{equation}
 X_{\ell+1} = Cov(X_{\ell},stride=2) 
\end{equation}
We propose AIL instead of those traditional sharpen transition layer. Suppose that
X$_{in}$ is the input of AIL with the size of (\emph{M},\emph{N},\emph{c}),  X$_{out}$ is the out of AIL with the size of (\emph{$\frac{M}{s}$},\emph{$\frac{N}{s}$},\emph{c'}), and the mathematical principle of AIL follow these equations:
\begin{equation}
 X = Relu(Cov(X_{in},1\times1))
\end{equation}

\begin{equation}
 W = Sigmoid(Cov(X_{in},3\times3))
\end{equation}
X in equation(3) extract the content matrix of the input with the operation RELU after Cov1$\times$1. W in equation(4) extract the attention matrix with the operation Sigmoid after Cov3$\times$3. Both of them have the size of (\emph{M},\emph{N},\emph{c'}).\\
Window shifting is one of the basic operation in convolutional networks, the operation is shown in Fig.~\ref{fig:pic20}. For mathematically intuitive, we are going to do the calculation on one of the selected window. Suppose that $\tilde{X}$,$\tilde{W}$ is one of the selected window from X,W.
\begin{equation}
 \tilde{X},\tilde{W} = Window(kernel\underline{\hspace{0.5em}}size=(m,n,c'),stride=s)(X,W)
\end{equation}

\begin{figure}[ht]
  \centering
  \includegraphics[width=1.0\linewidth]{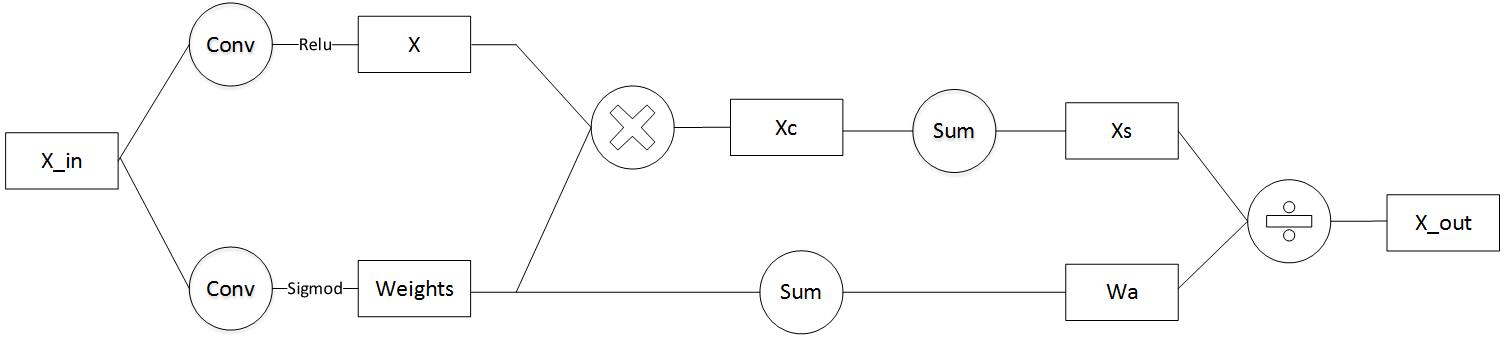}
  \caption{AIL computational graph.}
  \label{fig:pic4}
\end{figure}

Both of $\tilde{X}$,$\tilde{W}$ have the size of (\emph{m},\emph{n},\emph{c'}). Suppose that X$_{out}$ is the output of AIL with the shape of {(1,1,\emph{c'})}.
\begin{equation}
 \tilde{X}_{out,k} =\frac{\sum_{i}^{m}\sum_{j}^{n}(\tilde{W}_{i,j,k}\cdot \tilde{X}_{i,j,k})}{\sum_{i}^{m}\sum_{j}^{n}\tilde{W}_{i,j,k}+\epsilon}
\end{equation}
 $\tilde{X}_{out,k}$ stands for the value of $\tilde{X}_{out}$ in k$^{th}$ channel. $X_{out}$ equals to the stack by $\tilde{X}_{out}$ from all the sampling window in equation(5). $\epsilon$ is a small number equals to $10^{-8}$ to make sure the fraction won't divide by zero. Kernel size \emph{m},\emph{n} can arbitrary value during the feed-forward calculation. If the kernel \emph{m}=\emph{n}=constant then the calculation present as Conv+Pooling with attention mechanism, we name it as \emph{local attention incorporate layer}(LAIL). If \emph{m}=\emph{M} and \emph{n}=\emph{N}, there is only one window produced during the sampling, the window represent the whole image, the calculation extract the global information of the input image. We name it as \emph{global attention incorporate layer}(GAIL). For more concretely, the forward operation shows the following(Fig.~\ref{fig:pic4})  

\paragraph{Backward Pass.} $\tilde{X}$,$\tilde{W}$ in equation(5) have different representation of content and weight. These matrix learnt by the backward equation(7)(8):
\begin{equation}
\frac{\partial \tilde{X}_{out,k}}{\partial \tilde{X}_{x,y,k}}=\frac{\tilde{W}_{x,y,k}}{\sum_{i}^{m}\sum_{j}^{n}\tilde{W}_{i,j,k}+\epsilon}
\end{equation}

\begin{equation}
\frac{\partial \tilde{X}_{out,k}}{\partial \tilde{W}_{x,y,k}}=\frac{\sum_{i}^{m}\sum_{j}^{n}\tilde{W}_{i,j,k}\cdot(\tilde{X}_{x,y,k}-\tilde{X}_{i,j,k})}{(\sum_{i}^{m}\sum_{j}^{n}\tilde{W}_{i,j,k}^{2}+\epsilon)}
\end{equation}

$\frac{\partial \tilde{X}_{out,k}}{\partial \tilde{X}_{x,y,k}}$ stands for the gradient in the $\tilde{X}_{x,y,k}$ passed by$\tilde{X}_{out,k}$, it's pretty obvious that if the weight matrix W has a high value in point(x,y,k), the gradient of content X in this point will also have a high value. In global vision that if there are something important in this image, AIL will give a high gradient on those important area.

$\frac{\partial \tilde{X}_{out,k}}{\partial \tilde{W}_{x,y,k}}$ stands for the gradient in the $\tilde{W}_{x,y,k}$ passed by$\tilde{X}_{out,k}$, from the equation(8) we can know that AIL will likely to give a higher gradient to the point(x,y,k) if constant $\tilde{X}$ in this point has a high variance to other points. It's magically the same as human cognitive system that pay high attention on the very different things.

\subsection{Network Structure}

Recently DenseNet~\cite{Huang2016Densely} achieve a high accuracy in image classification problem. We use the \emph{Dense Block}~\cite{Huang2016Densely} as the basic block in AIN for the reason that DenseNet require substantially fewer parameters and less computation to achieve state-of-the-art performances. We implement AIN in the kaggle competition of \emph{Furniture-128}\footnote{\url{https://www.kaggle.com/c/imaterialist-challenge-furniture-2018}} which has various image size for training. The specific networks are shown in Table.~\ref{table1}

\begin{table}[h]
\renewcommand\arraystretch{1.5}
\centering
\caption{AIN architectures for  kaggle competition of \emph{Furniture-128}}
\label{table1}
\begin{tabular}{c|ccl}
\hline
layers         & \multicolumn{1}{c|}{Output Size} & \multicolumn{1}{c|}{AIN-121} & AIN-169 \\ \hline
Input          & \multicolumn{1}{c|}{M$\times$N$\times$3}        & \multicolumn{2}{c}{-}                 \\ \hline
Cov          & \multicolumn{1}{c|}{$\frac{M}{2}\times\frac{N}{2}\times64$}        & \multicolumn{2}{c}{kernel\underline{\hspace{0.5em}}size7$\times$7,stride2}                 \\ \hline
LAIL(1)        & \multicolumn{1}{c|}{$\frac{M}{4}\times\frac{N}{4}\times64$}   & \multicolumn{2}{c}{kernel\underline{\hspace{0.5em}}size3$\times$3,stride2}       \\ \hline
Dense Block(1) & \multicolumn{1}{c|}{$\frac{M}{4}\times\frac{N}{4}\times256$}   & \multicolumn{1}{c|}{$\binom{1\times 1 cov}{3\times 3 cov}\times6$}        & $\binom{1\times 1 cov}{3\times3 cov}\times6$ \\ \hline
LAIL(2)        & \multicolumn{1}{c|}{$\frac{M}{8}\times~\frac{N}{8}\times64$}   & \multicolumn{2}{c}{kernel\underline{\hspace{0.5em}}size3$\times$3,stride2}       \\ \hline
Dense Block(2) & \multicolumn{1}{c|}{$\frac{M}{8}\times~\frac{N}{8}\times256$}   & \multicolumn{1}{c|}{$\binom{1\times cov}{3\times 3 cov}\times12$}        &$\binom{1\times 1 cov}{3\times3 cov}\times12$         \\ \hline
LAIL(3)        & \multicolumn{1}{c|}{$\frac{M}{16}\times~\frac{N}{16}\times128$} & \multicolumn{2}{c}{kernel\underline{\hspace{0.5em}}size3$\times$3,stride2}        \\ \hline
Dense Block(3) & \multicolumn{1}{c|}{$\frac{M}{16}\times~\frac{N}{16}\times512$} & \multicolumn{1}{c|}{$\binom{1\times 1 cov}{3\times 3 cov}\times24$}        &$\binom{1\times 1 cov}{3\times 3 cov}\times32$         \\ \hline
LAIL(4)        & \multicolumn{1}{c|}{$\frac{M}{32}\times~\frac{N}{32}\times256$} & \multicolumn{2}{c}{kernel\underline{\hspace{0.5em}}size3$\times$3,stride2}        \\ \hline
Dense Block(4) & \multicolumn{1}{c|}{$\frac{M}{32}\times~\frac{N}{32}\times1024$} & \multicolumn{1}{c|}{$\binom{1\times 1 cov}{3\times 3 cov}\times16$}        &$\binom{1\times 1 cov}{3\times 3 cov}\times32$         \\ \hline
GAIL           & \multicolumn{1}{c|}{$1\times~1\times512$}         & \multicolumn{2}{c}{kernel\underline{\hspace{0.5em}}size$\frac{M}{32}\times~\frac{N}{32}$}        \\ \hline
FC,Softmax     & \multicolumn{3}{c}{128}                                                  \\ \hline                                               
\end{tabular}
\end{table}

\begin{figure}[ht]
  \begin{minipage}[t]{0.5\linewidth} 
  \centering
  \includegraphics[width=2.9in]{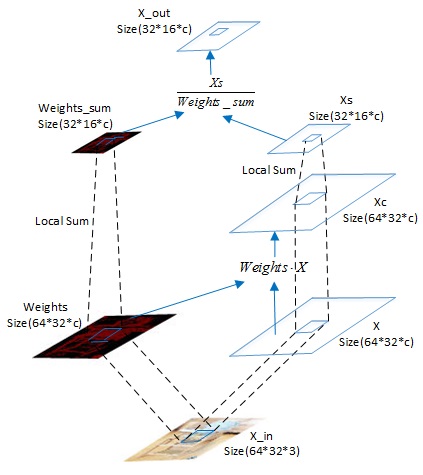}
  \caption{LAIL computational graph.}
  \label{fig:pic5}
  \end{minipage}%
  \begin{minipage}[t]{0.6\linewidth} 
  \centering
  \includegraphics[width=2.9in]{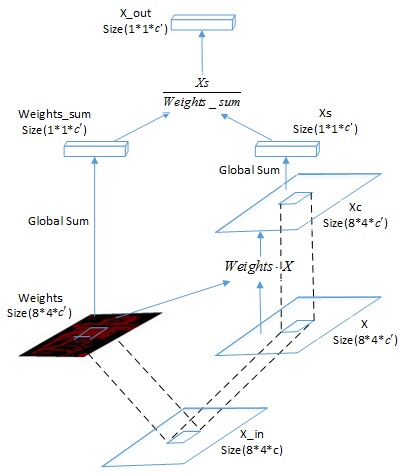}
  \caption{GAIL computational graph.}
  \label{fig:pic6}
  \end{minipage}%
\end{figure}

\paragraph{Local Attention Incorporate Layer.} The size of input tensors will decrease by \nicefrac{1}{s}~(s=stride) through LAIL, \emph{matrix Weights} produce by Cov+Sigmoid which gives a value between 0-1 as an attention gate, \emph{matrix X} produce by standard Cov+Relu which contains the key information of the image. The weighted average output is the sum of the \emph{matrix X} each weighted by \emph{matrix Weights}. A simple visualization example of LAIL is shown in Fig.~\ref{fig:pic5}.   

\paragraph{Global Attention Incorporate Layer.}
GAIL was designed to solve the arbitrary inputs problem. The output of GAIL will be a fixed size such as $1\times1\times512$ what ever the input size is. The computational logic of GAIL and LAIL are exactly the same. GAIL is the last layer of convolutional layers, the feed-forward feature will be highly abstract (e.g. Whether it is a object or the unimportant background). The image information will be transferred to the fixed number of channels to make sure that where will be no conflict between GAIL and Fully-Connected Layer. A visualization example of GAIL is shown in Fig.~\ref{fig:pic6}. 

\begin{figure}[h]
  \centering
  \includegraphics[width=1.0\linewidth]{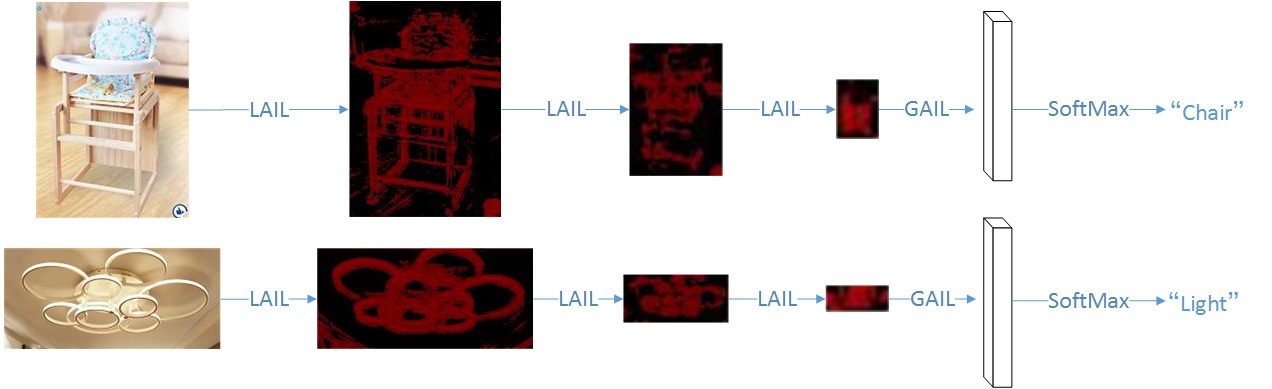}
  \caption{Visualization of AIN.}
  \label{fig:pic7}
\end{figure}

\paragraph{Implement Details.}
All image data set can use the network structure in Table.~\ref{table1} by tuning some model parameters (e.g. number of AIL, number of class). Experimentally we used AIN-121 to train the model in kaggle competition of \emph{Furniture-128}. Image size in the data set have various size. AIN-121 include 4$\times$LAIL, 4$\times$DenseBlock and 1$\times$GAIL before the Fully-Connected layer. All of the LAIL have the kernel\underline{\hspace{0.5em}}size of 3$\times$3. Dense Block are exactly the same as DenseNet~\cite{Huang2016Densely}. Part of the training data visualized in Fig.~\ref{fig:pic7}

\section{Experiments}
In this section we implement AIN in two applications: Image Classification and Audio Recognition on several data sets including CIFAR-10, Kaggle-\emph{Furniture128}, Speech Commands(tutorial data set in Tensorflow)\footnote{\url{https://www.tensorflow.org/versions/master/tutorials/audio_recognition}}. Given the limited computational resource and parameters, we compared the  performance of AIN to AlexNet and state-of-the-art architectures ResNet and DenseNet.  

\subsection{Image classification}
\paragraph{CIFAR10.}
The CIFAR-10 datasets consist of 60,000 with fixed color images size of 32$\times$32. 10 object classes with 50,000 training images and 10,000 test images. The broadly applied state-of-the-art
network structure DenseNet is used as baseline method. To make a fair comparison, we name the original dataset as "C10" without any data argumentation. Then we adopt a standard data augmentation scheme (mirroring/shifting), and keep most of the setting the same as in ResNet Paper~\cite{He2016Deep} and DenseNet Paper~\cite{Huang2016Densely}.  We
denote this data augmentation scheme by a “+” mark at the
end of the dataset name as "C10+".

We train all these models using \emph{SGD} optimizer with a mini-batch size of 64, set the initial learning rate to 0.1. The learning rate is divided by 10 at 150epoch, 225epoch. We terminate training at 300epoch. The main results on CIFAR10 are shown in Table.~\ref{table2}. To highlight general trends, we mark all results that outperform the existing state-of-the-art in \textbf{boldface} and
the overall best result in ${\color{blue} \textbf{blue}}$.

\begin{table}[h]
\renewcommand\arraystretch{1.2}
\centering
\caption{Error rates (\%) on CIFAR}
\label{table2}
\begin{tabular}{c|cc|cc}
\hline
Motheds           & Depth & Params & C10 & C10+ \\ \hline
AlexNet           & 8     & 4.5M   & 35.8& 26.4 \\ \hline
ResNet-110        & 110   & 1.7M       & -    & 6.43     \\ \hline
DenseNet(k=12)    & 40    & 1.04M   &7.00 & 5.36     \\
DenseNet(k=12)    & 100   & 7.08M   & 5.84    & \textbf{4.21}     \\ \hline
AIN(CNNBlock)     & 10    & 4.7M   &21.8 &14.2  \\ \hline
AIN(ResBlock)     & 110    & 2.31M   &-     &  6.26    \\\hline
AIN(DenseBlock)   & 40    & 1.57M   &6.72     & 5.15     \\
AIN(DenseBlock)   & 100   & 10.6M       &5.61     & ${\color{blue} \textbf{4.02}}$     \\ \hline
\end{tabular}
\end{table}

\paragraph{Furniture-128.}
The training dataset includes images from 128 furniture and home goods classes with various image size (e.g. 350$\times$350, 640$\times$480, 268$\times$400). It includes a total of 194,828 images for training and 6,400 images for validation and 12,800 images for testing. We do the image preprocess in two ways: The first way is to wrap the image into a fixed size 224$\times$224 associated with the standard data augmentation scheme (mirroring/shifting) just like the traditional way, we name this dataset as "furniture128+". The second way is to scale the image into the maximum slide size of 224(e.g.:350$\times$350 $\rightarrow$ 224$\times$224, 640$\times$480 $\rightarrow$ 224$\times$168, 268$\times$400 $\rightarrow$150$\times$224). Data augmentation scheme (mirroring/shifting) stays the same as the first way. The input data still remain the size distinctive, we name this dataset as "furniture128$^*$".

\begin{figure}[h]
  \begin{minipage}[t]{0.5\linewidth} 
  \centering
  \includegraphics[width=2.7in]{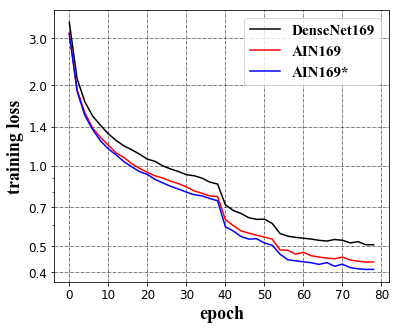}
  \caption{Loss and epochs}
  \label{fig:pic8}
  \end{minipage}%
  \begin{minipage}[t]{0.5\linewidth} 
  \centering
  \includegraphics[width=2.7in]{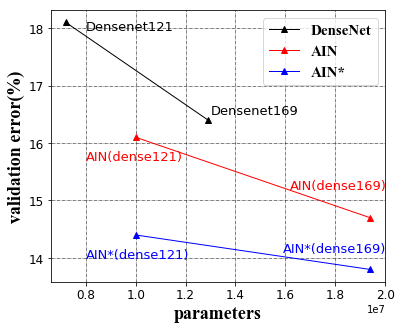}
  \caption{Error and parameters}
  \label{fig:pic9}
  \end{minipage}%
\end{figure}

We train all these models using \emph{adam} optimizer with a mini-batch size of 32, set the initial learning rate to 0.005. The learning rate is divided by 10 at 38epoch, 53epoch.  We terminate training at 80epoch. For DenseNet and Resnet, used the pre-trained model for initialization, with the input image scale into the fixed size. "furniture128$^*$" can only be trained in AIN cause of the various of image size. The main results on "Kaggle-furniture" are shown in Table.~\ref{table3}. The comparison between AIN and the base-line model in loss and error are shown in Fig.~\ref{fig:pic8} and Fig.~\ref{fig:pic9}, AIN$^*$ refers to the input images stay the various size. It's clearly to find that AIN$^*$ achieve a better performance in every side and AIN has a huge advantage in image classification problem especially for various input size scenario.

\begin{table}[h]
\renewcommand\arraystretch{1.2}
\centering
\caption{Error rates (\%) on furniture-128}
\label{table3}
\begin{tabular}{c|cc|c|c}
\hline
Motheds           & Depth & Params & furniture-128+ & furniture128$^*$\\ \hline
ResNet-101        & 101   & 42.9M      & 20.1          &-\\
ResNet-152        & 152   & 58.6M       &${\textbf{18.8}}$            &-\\ \hline
DenseNet121       & 121   & 7.2M  &18.1      &-\\
DenseNet169       & 169   & 12.9M &${\textbf{16.4}}$       & - \\ \hline
AIN(ResBlock)     & 101   & 44.1M      & 19.3          & 16.8  \\
AIN(ResBlock)     & 152   & 59.9M      & 18.2          & 16.5  \\ \hline
AIN(DenseBlock)   & 121   & 10.0M       & 16.1          & 14.4     \\
AIN(DenseBlock)   & 169   & 19.4M       &  14.7         & ${\color{blue} \textbf{13.8}}$    \\ \hline
\end{tabular}
\end{table}

\subsection{Audio recognition}
\paragraph{Speech Command.}
Speech Commands dataset, which consists of 65,000 wave audio files of people saying 30 different words. The length of the training wave is about 1 second, and there will be slightly different between every sample. We used 55,000 samples for training and 10,000 for testing. In the data preprocessing, we used \emph{"kaldi"} to extant the feature vector with FBANK algorithm, the setting of FBANK as fellow: (sampling frquency:16k, window length:25ms, window shift:10ms). Each frame have the feature vector of 40. Suppose that the length of wave is 1 second, this wave will transfer into a matrix of 100$\times$40 after the preprocessing.

We chose some widely used model for comparison. In order to make it fairly, we keep all setting mostly the same except the GAIL and other layers like LSTM GRU, The network structure with filters of 64 are shown in Table.~\ref{table4}. We train all these models using \emph{adam} with a mini-batch size of 64, set the initial learning rate to 0.001. The learning rate is divided by $\sqrt{10}$ if the loss stop improving for 5 epoch. We terminate training at 60epoch. The result shows in~Table.\ref{table5}

\begin{table}[h]
\begin{minipage}[t]{0.6\linewidth} 
\renewcommand\arraystretch{1.3}
\centering
\caption{Network structure of Speech Command}
\label{table4}
\begin{tabular}{c|cccc}
\hline
Layers        & \multicolumn{1}{c|}{Output Size} & \multicolumn{1}{c|}{AIN}  & \multicolumn{1}{c|}{LSTM} & GRU \\ \hline
Input         & \multicolumn{1}{c|}{$m\times$40}        & \multicolumn{3}{c}{-}                                       \\ \hline
Cov1D         & \multicolumn{1}{c|}{$\frac{m}{2}\times$64}     & \multicolumn{3}{c}{kernel\underline{\hspace{0.5em}}size15$\times$1,stride2}                     \\ \hline
TransferLyaer & \multicolumn{1}{c|}{$\frac{m}{4}\times$64}      & \multicolumn{1}{c|}{LAIL} & \multicolumn{2}{c}{maxpooling} \\ \hline
Cov1D         & \multicolumn{1}{c|}{$\frac{m}{4}\times$128}     & \multicolumn{3}{c}{kernel\underline{\hspace{0.5em}}size5$\times$1,stride1}                     \\ \hline
TransferLyaer & \multicolumn{1}{c|}{$1\times$128}       & \multicolumn{1}{c|}{GAIL} & \multicolumn{1}{c|}{lstm} & gru \\ \hline
FC,Softmax    & \multicolumn{4}{c}{30}                                                                        \\ \hline
\end{tabular}
\end{minipage}%
\begin{minipage}[t]{0.4\linewidth}
\renewcommand\arraystretch{1.3}
\centering
\caption{Error on Speech Command}
\label{table5}
\begin{tabular}{c|c|c}
\hline
Method    & Parms & Error(\%) \\ \hline
GRU\_64   & 172k  &  4.90         \\
GRU\_128  & 608k   & 4.73          \\ \hline
LSTM\_64  & 193k  & 4.80          \\
LSTM\_128 & 672k  & ${\textbf{4.34}}$          \\ \hline
AIN\_64   &156k   &3.61          \\
AIN\_128  &542k   &${\color{blue} \textbf{3.34}}$          \\ \hline
\end{tabular}
\end{minipage}%
\end{table}

\begin{figure}[h]
  \centering
  \includegraphics[width=2.8in]{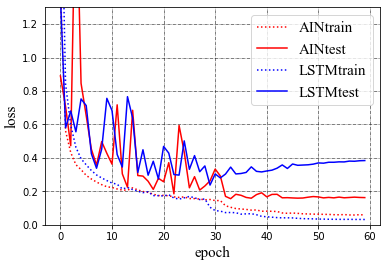}
  \caption{Loss between AIN$\_$128 and LSTM$\_$128.}
  \label{fig:pic12}
\end{figure}

Fig.~\ref{fig:pic12} shows that AIN has higher generalization ability than LSTM, AIN is less unlikely to overfit comparing to LSTM or GRU. AIN is a excellent neural-net structure for those end-to-end speech recognition problem with less parameters and higher accuracy. In the engineering field those CNN-based structures has higher potential to do data compression than sequence model.

\section{Discuss}
In this work, we introduce AIN which is a flexible solution for handling different scales, sizes, and aspect ratios used attention mechanism. These issues are important in computer vision because of the high similarity between AIN and human visual system. By visualizing the weight matrix in the LAIL (Fig.~\ref{fig:pic7}), we can know that the network have learnt the key information of the object. Thus it's not surprise to get a better result in CIFAR10 and Kaggle-\emph{Funiture128} and Speech Command than other model.
We believed that AIN can reach a higher accuracy by more detailed tuning of
hyper-parameters and learning rate schedules. There is still room for improvement in AIN. We are doing continuous improvement in our network and expect it to have a perfect performance in the future work.

\bibliographystyle{plain} 
\bibliography{}
\end{document}